\documentclass[10pt,twocolumn,letterpaper]{article}

\usepackage{cvpr}
\usepackage{times}
\usepackage{epsfig}
\usepackage{graphicx}
\usepackage{amsmath}
\usepackage{amssymb}
\usepackage{mathtools}
\usepackage{color}
\usepackage{xcolor}
\usepackage{url}            %
\usepackage{booktabs}       %
\usepackage{amsfonts}       %
\usepackage{nicefrac}       %
\usepackage{microtype}      %
\usepackage{subfig}
\usepackage{listings}
\usepackage{multirow}
\usepackage{makecell}
\usepackage{xspace}
\usepackage{mathabx}
\usepackage{stmaryrd}
\usepackage{bm}
\usepackage{enumitem}

\usepackage[square,numbers,sort,compress]{natbib}

\usepackage[pagebackref=true,breaklinks=true,letterpaper=true,colorlinks,bookmarks=false]{hyperref}
\usepackage{cleveref}

\usepackage{tikz}

\usepackage{pgfplots}

\usetikzlibrary{backgrounds,calc,chains,fit,matrix,positioning,shadows,shapes.misc,circuits.ee}
\tikzset{input/.style={}}
\tikzset{output/.style={}}
\tikzset{operator/.style={circle, draw, fill=white, minimum size=2.5ex, inner sep=0pt}}
\tikzset{filter/.style={rectangle, draw, fill=white, minimum size=3.5ex, inner xsep=1.5ex}}
\tikzset{other/.style={rounded rectangle, draw, fill=white, minimum size=3.5ex, inner xsep=1ex}}
\tikzset{branch/.style={circle, draw, fill=black, minimum size=.5ex, inner sep=0pt}}
\tikzset{rv/.style={circle, draw, thick, fill=white, minimum size=2.75ex, inner sep=0pt}}
\tikzset{ob/.style={circle, draw, thick, fill=lightgray, minimum size=2.75ex, inner sep=0pt}}
\tikzset{pa/.style={circle, draw, thick, fill=black, minimum size=1ex, inner sep=0pt}}
\tikzset{/tikz/thin/.style={line width=.9pt}}
\tikzset{/tikz/thick/.style={line width=1.4pt}}
\tikzset{every path/.style={thin}}
\tikzset{>=direction ee}

\pgfplotsset{compat=1.14}
\pgfplotsset{every axis/.append style={enlargelimits={abs=3pt},grid,axis lines=left}}
\pgfplotsset{every axis plot/.append style={thick,mark size=1.5pt,line join=bevel,mark options={solid}}}
\pgfplotsset{label style={font=\small}}
\pgfplotsset{tick label style={font=\footnotesize}}
\pgfplotsset{grid style={color=black!5}}
\pgfplotsset{legend style={draw=none,opacity=.85,font=\footnotesize,cells={anchor=west,opacity=1}}}
\pgfplotsset{every non boxed x axis/.style={xtick align=center,shorten <=-.5\pgflinewidth}}
\pgfplotsset{every non boxed y axis/.style={ytick align=center,shorten <=-.5\pgflinewidth}}
\pgfplotsset{every non boxed z axis/.style={ztick align=center,shorten <=-.5\pgflinewidth}}
\pgfplotsset{/pgf/number format/1000 sep={\,}}

\newcommand\R{\mathbb{R}}
\newcommand\ourcomb{$\bm{\dagger}$}

\newcommand{\orelu}{TLU}
\newcommand{\relu}{ReLU}
\newcommand{\ours}{[\textbf{Ours}]}

\newcommand{\papername}{FRN}
\newcommand{\batchnorm}{BN}
\newcommand{\momentnorm}{FRN}
\newcommand{\groupnorm}{GN}

\newcolumntype{L}[1]{>{\raggedright\let\newline\\\arraybackslash\hspace{0pt}}m{#1}}
\newcolumntype{C}[1]{>{\centering\let\newline\\\arraybackslash\hspace{0pt}}m{#1}}
\newcolumntype{R}[1]{>{\raggedleft\let\newline\\\arraybackslash\hspace{0pt}}m{#1}}

\definecolor{codegreen}{rgb}{0,0.6,0}
\definecolor{codegray}{rgb}{0.5,0.5,0.5}
\definecolor{codepurple}{rgb}{0.58,0,0.82}
\definecolor{backcolour}{rgb}{1.0,1.0,1.0}
 
\lstdefinestyle{mystyle}{
    float=[t],
    floatplacement=t,
    backgroundcolor=\color{backcolour},   
    commentstyle=\color{codegreen},
    keywordstyle=\color{magenta},
    numberstyle=\tiny\color{codegray},
    stringstyle=\color{codepurple},
    basicstyle=\ttfamily\scriptsize,
    breakatwhitespace=false,         
    breaklines=true,                 
    captionpos=t,                    
    keepspaces=true,                 
    numbersep=5pt,                  
    showspaces=false,                
    showstringspaces=false,
    showtabs=false,                  
    tabsize=2,
    frame=shadowbox,
    boxpos=t,
    aboveskip=0pt,
    belowskip=0pt
}
\cvprfinalcopy %

\def\cvprPaperID{1059} %

\begin{document}

\title{Filter Response Normalization Layer: Eliminating Batch Dependence in the Training of Deep Neural Networks}

\author{Saurabh Singh \qquad Shankar Krishnan\\
Google Research\\
{\tt\small \{saurabhsingh,skrishnan\}@google.com}
}

\maketitle

\begin{abstract}
Batch Normalization (BN) uses mini-batch statistics to normalize the activations during training, introducing dependence between mini-batch elements. This dependency can hurt the performance if the mini-batch size is too small, or if the elements are correlated. Several alternatives, such as Batch Renormalization and Group Normalization (GN), have been proposed to address this issue. However, they either do not match the performance of BN for large batches, or still exhibit degradation in performance for smaller batches, or introduce artificial constraints on the model architecture.
In this paper we propose the Filter Response Normalization (\papername{}) layer, a novel combination of a normalization and an activation function, that can be used as a replacement for other normalizations and activations. Our method operates on each activation channel of each batch element independently, eliminating the dependency on other batch elements. Our method outperforms BN and other alternatives in a variety of settings \textbf{for all batch sizes}. \papername{} layer performs $\approx \textbf{0.7-1.0}\%$ better than BN on top-1 validation accuracy with large mini-batch sizes for Imagenet classification using InceptionV3 and ResnetV2-50 architectures. Further, it performs $>~\textbf{1}\%$ better than GN on the same problem in the small mini-batch size regime. For object detection problem on COCO dataset, \papername{} layer outperforms all other methods by at least $\textbf{0.3-0.5}\%$ in all batch size regimes.

\end{abstract}

\vspace*{-0.1in}
\section{Introduction}
\label{sec:intro}

\begin{figure}[t]
\begin{center}
   \includegraphics[width=0.98\linewidth]{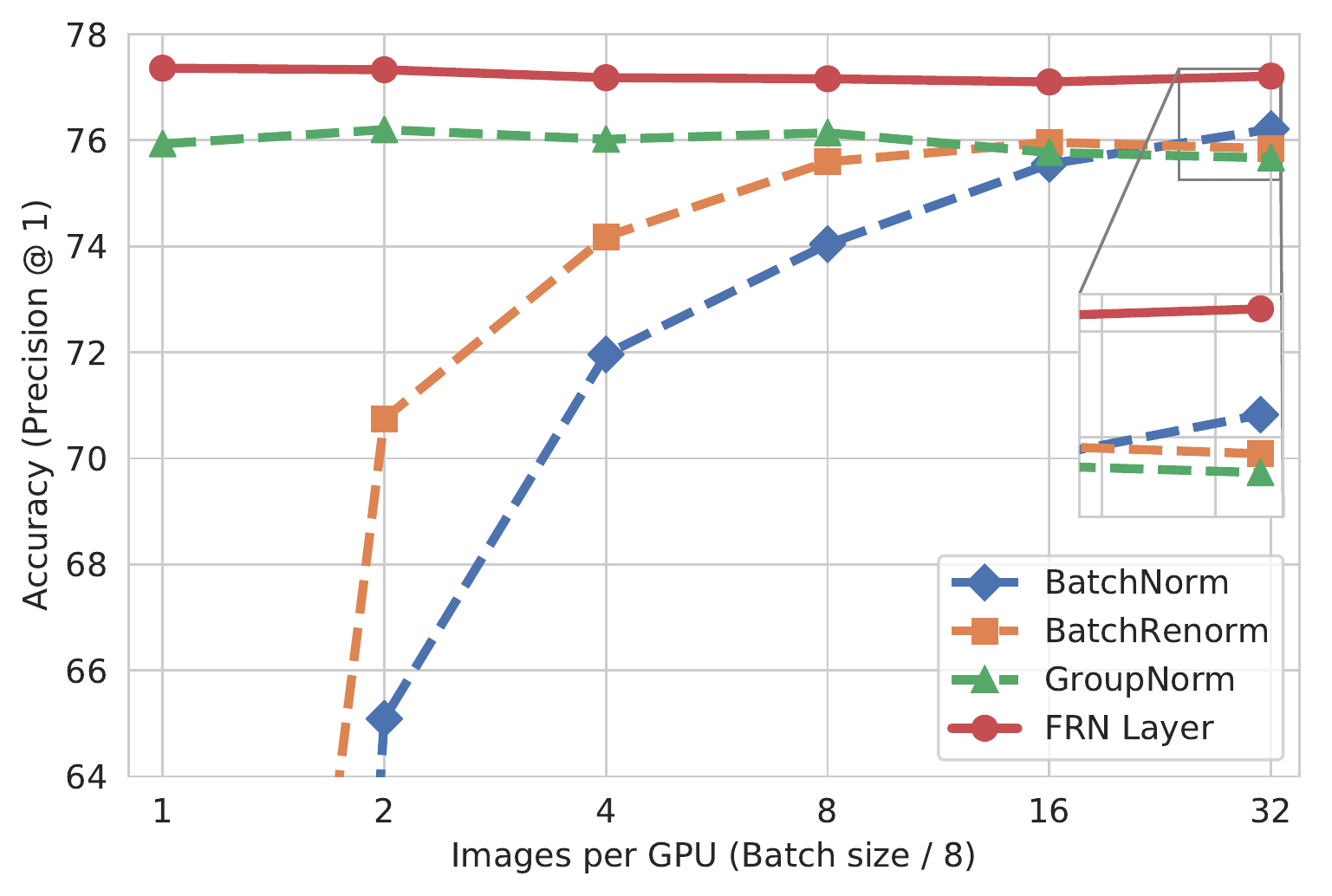}
\end{center}
\vspace{-0.5cm}
   \caption{Our method consistently outperforms other normalization methods, even at the largest batch size where other methods struggle in comparison to Batch Normalization (see inset). The figure reports the validation performance of ResNetV2-50 models trained using 8 GPUs with different batch sizes on ImageNet.}
\label{fig:acc_batchsize}
\vspace{-0.5cm}
\end{figure}

Batch normalization (BN)~\cite{batchnorm} is a cornerstone of current high performing deep neural network models. One often discussed drawback of \batchnorm{} is its reliance on sufficiently large batch sizes~\cite{batchrenorm, groupnorm, evalnorm}. When trained with small batch sizes, \batchnorm{} exhibits a significant degradation in performance.
This issue has been attributed to training and testing discrepancy arising from \batchnorm{}'s reliance on stochastic mini-batches~\cite{evalnorm}. As a result, several approaches have been proposed that ameliorate the issues due to stochasticity~\cite{batchrenorm, evalnorm} or offer alternatives~\cite{groupnorm, layernorm} by removing batch dependence.
However, these approaches don't match the performance of \batchnorm{} for large batch sizes (\Cref{fig:acc_batchsize}). Further, either they  still exhibit a degradation in performance for smaller batch sizes e.g. Batch Renormalization, or introduce constraints on the model architecture and size e.g. Group Normalization requires number of channels in a layer to be multiples of an ideal group size, such as 32.
In this work we propose \emph{Filter Response Normalization (\papername{}) layer}, consisting of a normalization and activation function, that eliminates these shortcomings altogether. Our method does not have any batch dependence, as it operates on each activation channel (filter response) of each batch sample independently, and outperforms \batchnorm{} and alternatives in a wide variety of evaluation settings. For example, in  \Cref{fig:acc_batchsize}, \papername{} layer outperforms other approaches by more than $1\%$ at \emph{all batch sizes} for ResNetV2-50 on ImageNet classification.

The reliance of \batchnorm{} on large batch sizes hinders the exploration of higher capacity models due to significantly higher memory requirements and imposes limitations on the performance of tasks that need to process larger inputs. For example, object detection and segmentation perform better with higher resolution inputs; similarly, video data tends to be very high dimensional. As a result, these systems are forced to trade-off between model capacity and ability to train with larger batch sizes. As evidenced in \Cref{fig:acc_batchsize}, our method addresses this by maintaining a consistent performance across a range of batch sizes.

\papername{} layer consists of two novel components that work together to yield high performance: 1) \emph{Filter Response Normalization (\papername{})}, a normalization method that independently normalizes the responses of each filter for each batch element by dividing them by the square root of their uncentered second moment, \emph{without performing any mean subtraction}, and 2) \emph{Thresholded Linear Unit (\orelu{})}, a pointwise activation that is parameterized by a learned rectification threshold, allowing for activations that are biased away from zero. \papername{} layer outperforms \batchnorm{} by more than $\textbf{0.7-1.0}\%$ with large mini-batch sizes for Imagenet classification using InceptionV3 and ResnetV2-50 architectures. Further, it performs $>~\textbf{1}\%$ better than GN on the same problem in the small mini-batch size regime. For object detection on COCO dataset, \papername{} layer outperforms all other methods by at least $\textbf{0.3-0.5}\%$ in all batch size regimes. Lastly, \papername{} layer maintains a consistent performance across all the batch sizes that we tested. In summary, \textbf{the proposed \papername{} layer does not rely on other batch elements or channels for normalization, yet outperforms \batchnorm{} and other alternatives for all batch sizes and in a variety of settings}.

\medskip
\noindent
\textbf{Contributions: }

\begin{enumerate}[leftmargin=*,noitemsep]
\item Filter Response Normalization (FRN), a normalization method that enables models trained with per-channel normalization to achieve high accuracy.
\item Thresholded Linear Unit (TLU), an activation function to use with FRN that further improves accuracy and outperforms \batchnorm{} at all batch sizes without any batch dependency. We refer to this combination as \papername{} layer.
\item Several insights and practical considerations that lead to the success of the combination of \momentnorm{} and \orelu{}.
\item A detailed experimental study comparing popular normalization methods on large scale image classification and object detection tasks using a variety of architectures.
\end{enumerate}

\section{Related work}

Normalization of training data has been known to aid in optimization. For example, whitening of inputs is a common practice for training shallow models such as Support Vector Machines and Logistic regression. Similarly, for training deep networks, normalization of inputs and intermediate representations has been recommended for efficient learning~\cite{lecun2012efficient,lecun1998efficient,glorot2010understanding}. 
Batch Normalization (BN)~\cite{batchnorm} accelerates learning and enables training of very deep neural network architectures by stabilizing the intermediate feature distributions. Stabilization is achieved by normalizing each activation channel independently, using the mean and variance statistics computed for that channel over the entire mini-batch. 
However, \batchnorm{} exhibits a dramatic degradation in performance when trained with smaller mini-batches~\cite{groupnorm, evalnorm}. Several approaches have been proposed to address this shortcoming, and can be grouped into two major categories: 1) Methods that reduce the train-test discrepancy in batch normalized models, 2) Sample based normalization methods that avoid batch normalization.

\medskip
\noindent
\textbf{Methods reducing train-test discrepancy in batch normalization.}
\citet{batchrenorm} notes that the discrepancy between the statistics that are used for normalization during training and testing may arise from the stochasticity due to small mini-batches and bias due to non-iid samples. They propose Batch Renormalization (BR) to reduce this discrepancy by constraining the mini-batch moments to a specific range, limiting the variation in mini-batch statistics during training. A key benefit of this approach is that the test time evaluation scheme of a model trained with BR is exactly the same as that for a model trained with \batchnorm{}. In comparison, EvalNorm~\cite{evalnorm} does not modify the training scheme. Instead, it proposes a correction to the normalization statistics for use only during evaluation. The major advantage of this method is that the model does not need to be retrained. However, both these methods still exhibit a degradation in performance for small mini-batches. Another approach is to engineer systems that can circumvent the issue by distributing larger batches across GPUs for tasks that require large inputs \cite{peng2017megdet}. However, this approach requires considerable GPU infrastructure.

\medskip
\noindent
\textbf{Methods avoiding normalization using mini-batches.}
Several approaches sidestep the issues encountered by \batchnorm{} by not relying on the stochastic mini-batch altogether~\cite{layernorm, groupnorm, instancenorm}. Instead, the normalization statistics are computed from the sample itself. Layer Normalization (LN)~\cite{layernorm} computes the normalization statistics from the entire layer i.e. using all the activation channels. In contrast, like \batchnorm{}, Instance Normalization (IN)~\cite{instancenorm} computes the normalization statistics for each channel independently, but only from the sample being normalized, as opposed to the entire batch, as \batchnorm{} does. IN was shown to be useful for style transfer applications, but was not successfully applied for recognition. Group Normalization (GN)~\cite{groupnorm} fills the middle ground between the two. It computes the normalization statistics over groups of channels. The ideal group size is experimentally determined. While, GN doesn't show  performance degradation for smaller batch sizes, it performs worse than \batchnorm{} for larger mini-batches (See \Cref{fig:acc_batchsize} here and Figure 1 in~\cite{groupnorm}). In addition, the size of groups required by GN imposes a constraint on the network size and architecture as every normalized layer needs to have number of channels that are a multiple of the ideal group size determined by GN.

\medskip
\noindent
\textbf{Other approaches.} Weight Normalization~\cite{salimans2016weight} proposes a reparameterization of the filters in terms of a direction and a scale and reports accelerated convergence. Normalization Propagation~\cite{arpit2016normalization} uses idealized moment estimates to normalize every layer. Refer to \citet{ren2016normalizing} for a unifying view of various normalization approaches.  Divisive normalization (DN)~\cite{bonds1989role,heeger1992normalization} has been proposed to normalize each activation with a function of the neighboring activations and has been studied in a variety of contexts including density modeling~\cite{balle2016density}, image compression~\cite{balle2017end, balle2018variational}, sensitivity maximization~\cite{carandini2012normalization}, distributed neural representation~\cite{simoncelli1998model} and attention~\cite{reynolds2009normalization}, among others.

\begin{figure}[t]
\resizebox{\linewidth}{!}{
  \begin{tikzpicture}[x=1em,y=1em]
    \node [input] (x) {$\bm{x}$};
    \node [other] (frn) at ($(x)+(7.5,0)$) {\begin{tabular}{c}$\nu^2=\sum_i x_i^2/N$ \\ $y_i = \gamma\frac{x_i}{\sqrt{\nu^2 + \epsilon}} + \beta$ \end{tabular}};
    \node [other] (tlu) at ($(frn)+(11,0)$) {$z_i = \max(y_i, \tau)$};
    \node [output] (z) at ($(tlu)+(6.5,0)$) {$\bm{z}$};
    
    \node [below=1.8em] at (frn) {FRN};
    \node [below=1.8em] at (tlu) {TLU};

    \draw[->] (x) -- (frn);
    \draw[->] (frn) -- (tlu) node[midway,above] {$\bm{y}$};
    \draw[->] (tlu) -- (z);

    \begin{pgfonlayer}{background}
      \node[fill=black!8,rounded corners=3ex,draw,thick,fit=(frn)(tlu),inner xsep=2ex,inner ysep=3.5ex] (frn_layer) {};
    \end{pgfonlayer}
    \node [above,text height=1.5ex, text depth=.25ex] at (frn_layer.north) {FRN Layer};
  \end{tikzpicture}\hfill
  }
  \caption{A schematic of the proposed FRN Layer.}
  \label{fig:FRNlayer}
  \vspace{-0.5cm}
\end{figure}
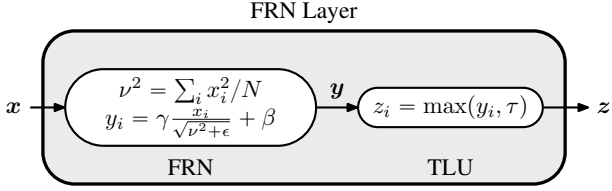

\section{Approach}

Our goal is to eliminate the batch dependency in the training of deep neural networks without sacrificing the performance gains of \batchnorm{} at large batch sizes. We start this section with the main details of our proposal and follow that with a discussion of the rationale behind our approach.

\subsection{Filter Response Normalization with Thresholded Activation}
\label{sec:frn}

We will assume for the purpose of exposition that we are dealing with the feed-forward convolutional neural network. We follow the usual convention that the filter responses (activation maps) produced after a convolution operation are a 4D tensor $\bm{X}$ with shape $[B, W, H, C]$, where $B$ is the mini-batch size, $W, H$ are the spatial extents of the map, and $C$ is the number of filters used in convolution. $C$ is also referred to as output channels. Let $\bm{x} = \bm{X}_{b,:,:,c} \in \R^{N}$, where $N = W \times H$, be the vector of filter responses for the $c^{th}$ filter for the $b^{th}$ batch point. 
Let $\nu^2 = \sum_i x_i^2/N$, be the mean squared norm of $\bm{x}$. Then we propose Filter Response Normalization (\momentnorm{}) as following:
\begin{align}
\label{eq:chan_norm}
\hat{\bm{x}} = \frac{\bm{x}}{\sqrt{\nu^2 + \epsilon}},
\end{align}
where $\epsilon$ is a small positive constant to prevent division by zero. A few observations are in order about the proposed normalization scheme:

\begin{enumerate}[leftmargin=*,noitemsep]
    \item Similar to other normalization schemes, FRN removes the scaling due to both the filter weights and pre-activations. This is known~\cite{salimans2016weight} to remove noisy updates along the direction of the weights and reduce gradient covariance.
    \item One of the main differences in our proposal is that we do not remove the mean prior to normalization. While mean subtraction was an important aspect of Batch Normalization, it is arbitrary and without real justification for normalization schemes that are batch independent.
    \item Our normalization is done on a \emph{per-channel} basis. This ensures that all filters (or rows of a weight matrix) have the same relative importance in the final model.
    \item At a first glance, FRN appears very similar to Local Response Normalization (LRN) proposed in ~\citet{Alexnet2012}. However, among other differences, LRN does normalization over adjacent channels at the same spatial location, while FRN is a global normalization over the spatial extent.
\end{enumerate}

As with other schemes, we also perform an affine transform after normalization so that the network can undo the effects of the normalization:
\begin{align}
\label{eq:affine_xform}
\bm{y} = \gamma \hat{\bm{x}} + \beta,
\end{align}
where $\gamma$ and $\beta$ are learned parameters. The final addition to our \papername{} layer is the activation function. 

\subsubsection{Thresholded Linear Unit (\orelu{})}
\label{sec:offsetrelu}

Lack of mean centering in FRN can lead to activations having an arbitrary bias away from zero. Such a bias in conjunction with ReLU can have a detrimental effect on learning and lead to poor performance and dead units. We propose to address this issue by augmenting ReLU with a learned threshold $\tau$ to yield \orelu{} defined as:
\begin{align}
\label{eq:offsetrelu}
\bm{z} = \max(\bm{y}, \tau)
\end{align}
Since $\max(\bm{y}, \tau){=}\max(\bm{y}-\tau,0){+}\tau{=}\relu{(\bm{y}{-}\tau)}{+}\tau$, the effect of \orelu{} activation is the same as having a shared bias before and after \relu{}. However, this does not appear to be identical to absorbing the biases in the previous and subsequent layers based on our experiments. We hypothesize that the form of \orelu{} is more favorable for optimization. \orelu{} significantly improves the performance of models using \momentnorm{} (see \Cref{tab:compare_tlu}), outperforming \batchnorm{} and other alternatives, and leads to our method, \papername{} layer. Figure~\ref{fig:FRNlayer} shows the schematic for our proposed \papername{} layer.

\subsection{Gradients of \papername{} Layer}
\label{sec:gradients}

In this section, we derive expressions for the gradients that flow through the network in the presence of the \papername{} layer. Since all the transformations are performed channel-wise, we only derive the per-channel gradients. Let us assume that somewhere in the network, the activations $\bm{x}$ are fed to the \papername{} layer and the output is $\bm{z}$ (following the transformations described in equations~(\ref{eq:chan_norm}), ~(\ref{eq:affine_xform}), and ~(\ref{eq:offsetrelu})). Let $f(\bm{z})$ be the mapping that the network applies to $\bm{z}$, with gradients $\frac{\partial f}{\partial \bm z}$ flowing backwards. Note that the parameters $\gamma$, $\beta$ and $\tau$ are vectors of size \emph{num\_channels}, and so the per channel updates are scalar.
\begin{align}
\label{eq:orelu_grad}
\frac{\partial z_i}{\partial \tau} = 
\begin{cases}
0,& \text{if } y_i \geq \tau \\
1,& \text{otherwise}
\end{cases}
\end{align}
Note that the gradients $\frac{\partial z_i}{\partial y_i}$ are just the same as above, but with the cases reversed. Then the gradient update to $\tau$ is of the form
\begin{align}
\label{eq:tau_grad}
\frac{\partial f}{\partial \tau} = \sum_{b=1}^B \left( \frac{\partial f}{\partial \bm{z}_b} \right)^T \frac{\partial \bm{z}_b}{\partial \tau},
\end{align}
where $\bm{z}_b$ is the vector of per-channel activations of the $b^{th}$ batch point. Gradients w.r.t $\gamma$ and $\beta$ are as follows:
\begin{align}
\label{eq:gamma_beta_grad}
\left( \frac{\partial f}{\partial \gamma}, \frac{\partial f}{\partial \beta} \right) &= \left( \sum_{b=1}^B \frac{\partial f^T}{\partial \bm{y}_b} \hat{\bm{x}_b}, \sum_{b=1}^B \frac{\partial f}{\partial \bm{y}_b} \right)
\end{align}
Using eqn.~(\ref{eq:affine_xform}), we can see that $\frac{\partial f}{\partial \hat{\bm{x}}}{=}\gamma \frac{\partial f}{\partial \bm{y}}$. Finally, the gradients that flow back from the \papername{} layer can be written as
\begin{align}
\label{eq:x_grad}
\frac{\partial f}{\partial \bm{x}} &= \frac{1}{\sqrt{\nu^2 + \epsilon}} \left( I - \frac{\hat{\bm{x}} \hat{\bm{x}}^T}{N} \right) \frac{\partial f}{\partial \hat{\bm{x}}}
\end{align}
We make a couple of observations about the gradients. Eqn.~(\ref{eq:tau_grad}) suggests that part of the gradients that get suppressed in a regular ReLU activation are now used to update $\tau$, and in some sense are \emph{not wasted}. Since $||\hat{\bm{x}}||_2^2 = N$,  eqn.~(\ref{eq:x_grad}) shows that the gradients w.r.t to $\bm{x}$ are orthogonal to $\bm{x}$ (provided $\epsilon$ = 0) because $(I - \hat{\bm{x}} \hat{\bm{x}}^T/N)$ projects out the component in the direction of $\hat{\bm{x}}$. This property is not unique to our normalization, but is known to help in reducing variance of gradients during SGD and benefit optimization~\cite{salimans2016weight}.

\subsection{Learning $\bm{\epsilon}$}
\label{sec:learned_eps}

\begin{figure}[t]
\begin{center}
   \includegraphics[width=0.8\linewidth]{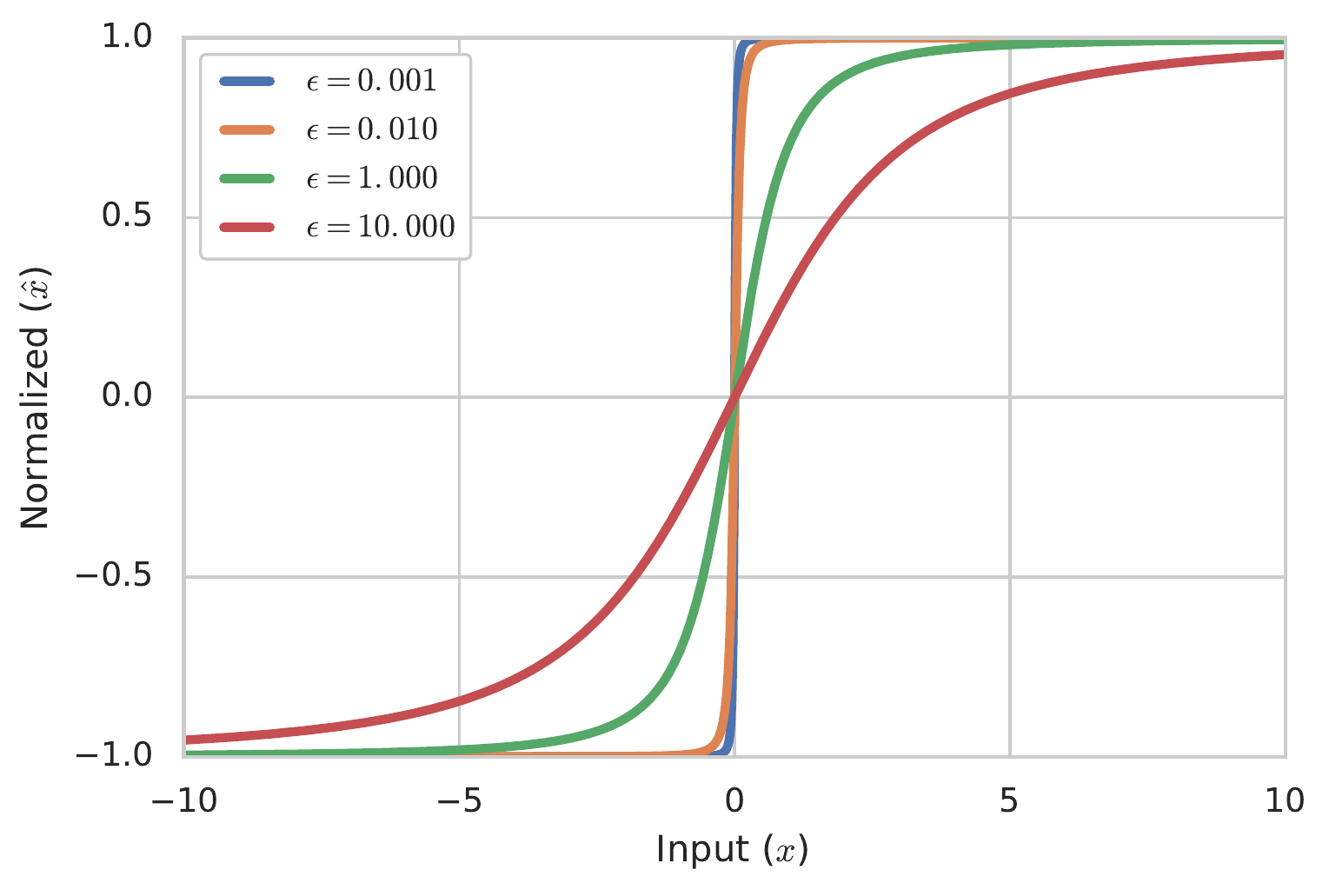}
\end{center}
\vspace{-0.6cm}
   \caption{Effect of $\epsilon$ on normalized activations for the case of $N=1$. For very small values of $\epsilon$, \momentnorm{} turns into a step function while for higher values it behaves like a softsign function, allowing the gradients to flow. Having a learnable epsilon is crucial in models with fully connected layers or low-dimensional activation maps.}
\label{fig:lambdaeffect}
\vspace{-0.4cm}
\end{figure}

In our discussion so far, we have assumed that the filter responses have a large spatial extent of size $N = W{\times}H$. However, there are situations in real networks like InceptionV3~\cite{inceptionv3} and VGG-A~\cite{vggnet}, where some layers produce $1 \times 1$ activation maps. In this setting ($N{=}1$), for small value of $\epsilon$, the proposed normalization as in \Cref{eq:chan_norm} turns into a sign function (see \Cref{fig:lambdaeffect}), and has very small gradients almost everywhere. This will invariable affect the learning adversely. In contrast, higher values of $\epsilon$ lead to variants of smoother soft sign function that are more amenable to learning. Therefore, appropriate value of $\epsilon$ becomes crucial for models that are fully connected or lead to $1{\times}1$ activation maps. Empirically, we turn $\epsilon$ into a learnable parameter (initialized at $10^{-4}$) for such models. For other models, we use a fixed constant value of $10^{-6}$. In our experiments, we show that the learnable parameterization is useful for training InceptionV3 model where the Auxiliary logits head produces $1{\times}1$ activation maps, and for the VGG-A~\cite{vggnet} architecture that uses fully connected layers.

\medskip
\noindent
Since $\epsilon > 0$, we explored two alternative parameterizations to enforce this constraint: absolute value and exponential. While both trained well, the absolute value parameterization $\epsilon{=}10^{-6}{+}|\epsilon_l|$ ($\epsilon_l$ being a learned parameter), produced consistently better empirical results. Parameterizations of this form are also preferable because the gradient magnitudes for $\epsilon_l$ are independent of the value of $\epsilon$.

\subsection{Mean Centering}
\label{sec:mean_centering}

Batch Normalization was proposed to counter the effects of \emph{internal covariate shift} during training of a deep neural network. The solution was to keep the statistics of distribution of activations over the data set invariant; and as a practical matter, they choose to normalize the first and second moments of mini-batch at each step. Batch independent alternatives that include mean centering are not justified by any particular consideration, and appear merely to be a legacy of Batch Normalization.

Consider the example of Instance Normalization (IN). Using the same notation as \Cref{sec:frn}, IN computes the  normalized activations using the channel statistics $\mu{=}\sum_i x_i/N$ and $\sigma^2{=}\sum_i (x_i{-}\mu)^2/N$ as following:
\begin{align}
\hat{\bm{x}} = \frac{\bm{x} - \mu}{\sqrt{\sigma^2 + \epsilon}}
\label{eq:instancenorm}
\end{align}
As the size of the activation map decreases (as is common in the layers closer to the output which are subject to downsampling, or due to the presence of fully connected layers), IN produces zero activations. Layer and Group Normalization are ways to circumvent this issue by normalizing across (all or subset of) channels. Since individual filters are responsible for each channel's activations, normalizing across channels introduces unnecessary interactions in the filter updates. Hence, it appears that the only principled approach is to normalize each channel of the activation map separately without resorting to mean centering. This also has the desirable effect of removing the relative scaling between filters, which is known to greatly aid in optimization.

A negative impact of not performing mean centering is that the activations can be biased arbitrarily away from zero, rendering ReLU activation less than ideal. We mitigate this issue by introducing the Thresholded Linear Unit (TLU) in \Cref{sec:offsetrelu}. Empirically, the combination of uncentered normalization with the TLU activation outperforms \batchnorm{} and all other alternatives.

\subsection{Implementation and Practical Considerations}
\label{sec:learning_rate}

\papername{} is easy to implement in automatic differentiation frameworks. We provide an example implementation using the python API for Tensorflow in \Cref{lst:implementation}. Further, for FRN to achieve peak accuracy, we found that care must be exercised with the following practical considerations. Sensitivity of FRN to each of these is architecture dependent.

\medskip
\noindent
\textbf{Learning Rate Schedule:} We found that the more common step decay learning rate (LR) schedule was not optimal for FRN. Instead, continuous LR decay schedules such as cosine decay (without restarts) performed better for all methods and eliminated the need to tune step decay hyper parameters. 

\medskip
\noindent
\textbf{Warm-up:}
Since \papername{} does not perform mean centering, we empirically found that certain architectures are more sensitive to the choice of initial LR. Setting a high initial LR causes large updates, that lead to large activations in the early part of the training and result in a slowdown in the learning. This is due to the $\frac{1}{\sqrt{\nu^2 + \epsilon}}$ factor in the gradient of $\frac{\partial f}{\partial \bm{x}}$ (see \Cref{eq:x_grad}). This happens more often in architectures that employ several max pooling layers, like VGG-A. We address this by using an initial warp-up phase where the LR is slowly increased from 0 to the peak value. Since all our experiments use cosine LR decay schedule, we use a cosine warm-up schedule as well. Note that a warmup phase is quite common and  frequently used in training ~\cite{resnets, resnetsv2,Imagenet2017}.

\medskip
\noindent
\textbf{Learnable $\bm{\epsilon}$:}
As discussed in \Cref{sec:learned_eps}, for models that use $1{\times}1$ activation maps, it is crucial to turn the $\epsilon$ into a learned parameter and initialize with a larger value to prevent step function like behavior and enable training.

\begin{figure}[t]
\vspace{-1mm}
\begin{lstlisting}[language=Python,caption=Tensorflow implementation of \papername{} layer,label=lst:implementation]
def FRNLayer(x, tau, beta, gamma, eps=1e-6):
  # x: Input tensor of shape [BxHxWxC].
  # tau, beta, gamma: Variables of shape [1, 1, 1, C].
  # eps: A scalar constant or learnable variable.
  
  # Compute the mean norm of activations per channel.
  nu2 = tf.reduce_mean(tf.square(x), axis=[1, 2], keepdims=True)
  
  # Perform FRN.
  x = x * tf.rsqrt(nu2 + tf.abs(eps))
  
  # Return after applying the Offset-ReLU non-linearity.
  return tf.maximum(gamma * x + beta, tau)
\end{lstlisting}
\vspace{-0.5cm}
\end{figure}

\section{Experiments}
We evaluate our method extensively on two tasks: 1) Image classification on Imagenet, and 2) Object detection on COCO. While Image classification is the de-facto standard for evaluation, Object detection typically requires high resolution inputs and is particularly constrained by the large batch size requirement of \batchnorm{}. On Imagenet classification, we show that our method outperforms other normalization methods on three different network architectures. Further, our method does this consistently at all batch sizes we experimented with. Finally, we validate the performance of our method on Object Detection where it outperforms other normalization methods on all batch sizes as well.

\subsection{ImageNet Classification} \label{sec:imagenet}

\noindent
\textbf{Dataset:} ImageNet classification dataset~\cite{imagenet} consists  of 1000 classes. We train on the ${\sim}1.28$M training images and report results on the 50000 validation images. For all models in this section, we resize the images to $299{\times}299$ and use data augmentation from~\cite{szegedy2017inception} at training time.

\medskip
\noindent
\textbf{Model architectures:} We provide comparisons using three different model architectures: 1) ResnetV2-50~\cite{resnetsv2}: Popular model with identity shortcuts, 2) InceptionV3~\cite{szegedy2016rethinking}: High performing model \emph{without} identity shortcuts and fully connected layers and, 3) VGG-A~\cite{vggnet}: Feed forward model with a mix of convolutional and fully connected layers. For all models using GN, we use a group size of 32. However, since VGG-A does not use a multiple of 32 filters in all layers, we increase the number of filters to nearest multiple.

\medskip
\noindent
\textbf{Training:}
We follow the training setup used by \citet{resnets}. All models are trained using synchronous SGD across 8 GPUs for 300000 steps. Gradients are computed by averaging across all GPUs. For BN, the normalization statistics are computed per GPU. This setup is common for multi-GPU training using synchronous SGD in Tensorflow and PyTorch.
An initial learning rate of $0.1{\times}\texttt{batch\_size}/256$ and cosine decay schedule is used. We follow~\cite{resnets, resnetsv2} for other implementation details. Results are reported using two image classification metrics: 1) `Precision@1' measures the accuracy using the highest scoring class (top-1 prediction) while, 2) `Recall@5' measures the accuracy using top-5 scoring classes.

\begin{table}[t]
    \centering
    \caption{\textbf{\papername{} layer outperforms \batchnorm{} and other normalization methods for large batch size} on Imagenet Classification for ResnetV2-50~\cite{resnetsv2} and InceptionV3~\cite{szegedy2016rethinking}.}
    \resizebox{0.8\linewidth}{!}{
    \begin{tabular}{@{}l c c c c@{}}
    \toprule
    \multirow{2}{*}[-2pt]{Method} & \multicolumn{2}{@{}c@{}}{ResnetV2 50} & \multicolumn{2}{@{}c@{}}{InceptionV3} \\
    \cmidrule(lr{0.25em}){2-3} \cmidrule(lr{0.25em}){4-5}
    & P@1 & R@5 & P@1 & R@5 \\
    \midrule
    Batchnorm & 76.21 & 92.98 & 78.24 & 94.07 \\
    BatchRenorm & 75.85 & 92.90 & 78.19 & 94.01 \\
    Groupnorm & 75.67 & 92.70 & 78.14 & 93.98 \\
    Layernorm & 72.75 & 91.19 & 76.75 & 93.37 \\
    Instancenorm & 71.63 & 90.46 & 73.93 & 91.60 \\
    \papername{} layer \ours  & \textbf{77.21} & \textbf{93.57} & \textbf{78.95} & \textbf{94.49} \\
    \bottomrule
    \end{tabular}
    }
    \label{tab:imagenetnormalization}
\end{table}

\begin{table}[t]
  \caption{\textbf{Effect of mini-batch size used for normalization} on ImageNet classification for ResnetV2-50~\cite{resnetsv2}.}
  \vspace{-0.2cm}
  \label{tab:imagenet_batchsizes}
  \centering
  \renewcommand{\arraystretch}{1.0}
  \renewcommand{\tabcolsep}{1.3mm}
  \resizebox{\linewidth}{!}{
  \begin{tabular}{@{}L{0.25cm} L{2.6cm} c c c c c c c@{}}
    \toprule
    \multicolumn{2}{r}{\small Images per GPU $\rightarrow$} & 32 & 16 & 8 & 4 & 2 & 1 \\
    \midrule
    \multirow{5}{*}[-2pt]{\rotatebox{90}{Precision@1}} 
                   & Batchnorm & 76.21 & 75.55 & 74.04 & 71.96 & 65.09 & 1.58 \\
                   & Renorm  & 75.85 & 75.96 & 75.59 & 74.18 & 70.75 & 37.55 \\
    \cmidrule(lr{0.25em}){2-8}
                   & Groupnorm & 75.67 & 75.77 & 76.14 & 76.02 & 76.20 & 75.93 \\
                   & \papername{} layer \ours & 77.21 & 77.10 & 77.16 & 77.18 & 77.33 & 77.36 \\
                   & $\Delta$ & \textbf{+1.54} & \textbf{+1.33} & \textbf{+1.02} & \textbf{+1.16} & \textbf{+1.13} & \textbf{+1.43} \\
    \midrule
    \multirow{5}{*}[-2pt]{\rotatebox{90}{Recall@5}} 
                   & Batchnorm & 92.98 & 92.81 & 92.12 & 90.98 & 86.51 & 4.00 \\
                   & Renorm & 92.90 & 92.98 & 92.80 & 92.10 & 89.81 & 57.18 \\
    \cmidrule(lr{0.25em}){2-8}
                   & Groupnorm & 92.70 & 92.72 & 92.89 & 92.87 & 92.92 & 92.73 \\
                   & \papername{} layer \ours & 93.62 & 93.59 & 93.60 & 93.49 & 93.61 & 93.61 \\
                   & $\Delta$ & \textbf{+0.92} & \textbf{+0.87} & \textbf{+0.71} & \textbf{+0.62} & \textbf{+0.69}  & \textbf{+0.88} \\
    \bottomrule
  \end{tabular}
  }
  \vspace{-0.2cm}
\end{table}

\begin{figure}[t]
\begin{center}
   \includegraphics[width=0.98\linewidth]{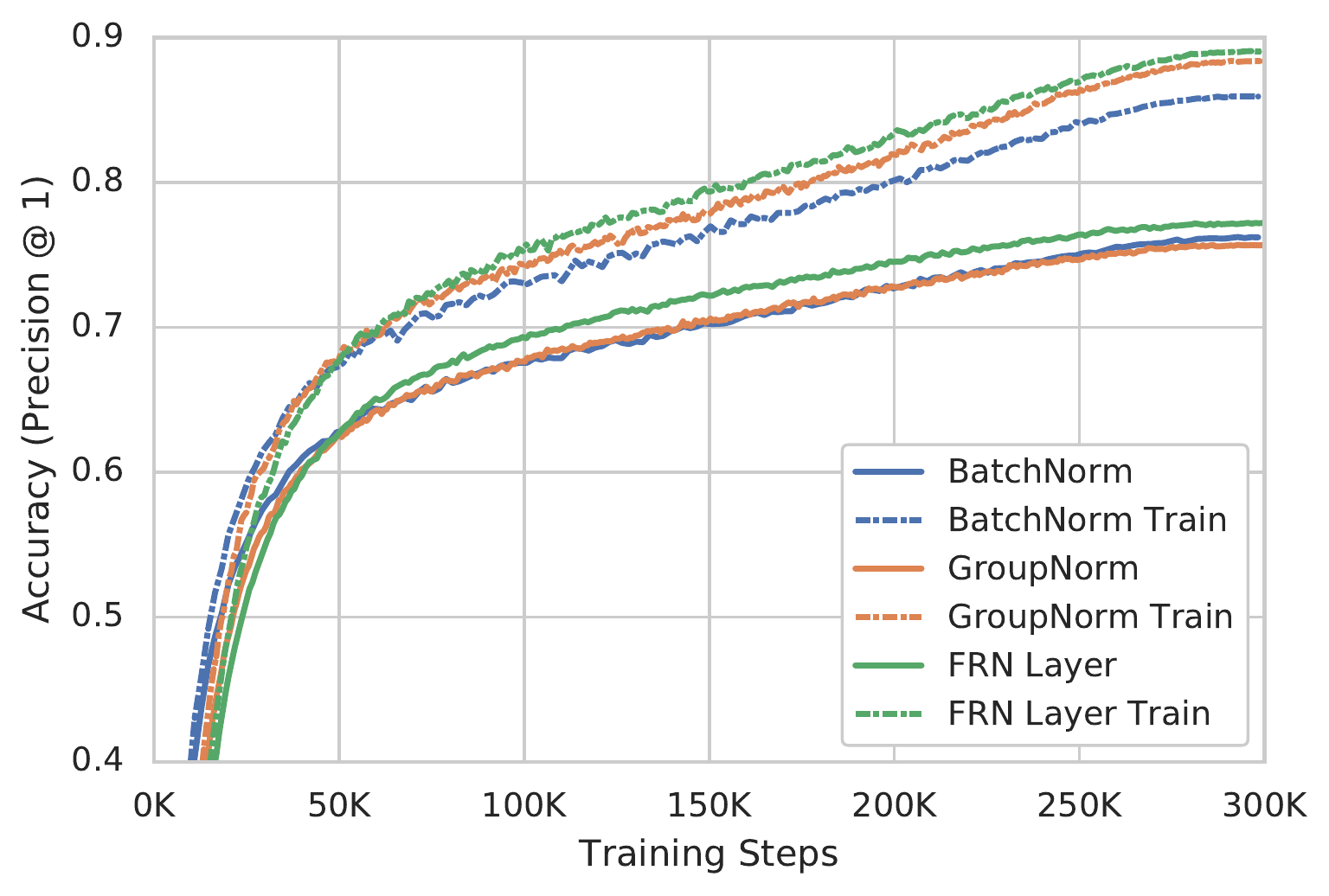}
\end{center}
\vspace{-0.7cm}
   \caption{\textbf{Comparison of training and validation curves} of various normalization methods for Imagenet Classification using ResnetV2-50 model.}
\label{fig:train_test_curves}
\vspace{-0.4cm}
\end{figure}

\medskip
\noindent
\textbf{Comparison with normalization methods:} In \Cref{tab:imagenetnormalization} we compare our method with various normalization methods for the regular batch size of 32 images/GPU. This results in an effective batch size of $32 \times 8 = 256$ and is the most favorable configuration for \batchnorm{}. This is the strongest baseline for image classification and all the alternatives to \batchnorm{} have struggled in this setting, underperforming \batchnorm{}. \emph{Even for this large batch size, \papername{} outperforms all the methods, including \batchnorm{},} with a healthy margin on both the architectures. Key takeaway is that \emph{batch dependent training is not necessary for high performance}. At this large batch size, the next best performing normalization schemes are \batchnorm{} and BatchRenorm, both of which are batch normalized methods, followed by other sample based normalization methods. \Cref{fig:train_test_curves} compares the training and validation 'Precision@1' curves for various normalization methods using the ResnetV2-50 architecture. We observe that \papername{} layer achieves both higher training and validation accuracies than BN indicating that removal of stochastic batch dependence eases \emph{optimization} allowing model to train better. The generalization gap, i.e. difference between training and validation accuracy, has also increased. However, improved optimization results in a net performance gain on validation. In comparison, GN also achieves lower training error than BN but performs worse on validation.

\medskip
\noindent
\textbf{Effect of small number of images per GPU:}
We study the impact of mini-batch sizes used for normalization (images/GPU) on the performance of various methods in \Cref{fig:acc_batchsize} and \Cref{tab:imagenet_batchsizes}. All methods are trained with 8 GPUs using six different total batch sizes of 8, 16, 32, 64, 128, 256, divided into equal number of images per GPU leading to 1, 2, 4, 8, 16, and 32 images/GPU. \batchnorm{} is known to degrade in performance when the batch size is small~\cite{batchrenorm, evalnorm} as evident in \Cref{fig:acc_batchsize}. GroupNorm (\groupnorm{}) exhibits a more consistent performance underperforming \batchnorm{} only at the largest batch size. Batch renormalization outperforms \groupnorm{} at the largest two batch sizes but shows a degradation in performance for the smaller batch sizes. \emph{Our method, \papername{}, consistently outperforms all the normalization methods at all batch sizes}. 

\medskip
\noindent
\textbf{Analyzing the effect of \momentnorm{} and \orelu{}:}
In \Cref{tab:analyzing_mn_tlu} we perform a detailed ablation study of the effect of \momentnorm{} and \orelu{}. We combine them with various normalization methods -- BatchNorm (BN), GroupNorm (GN), LayerNorm (LN) and InstanceNorm (IN), and train models for each combination for two high performing, but different, model architectures -- ResnetV2-50 and InceptionV3. 
We either replace \relu{} activation with \orelu{}, or modify the normalization technique to suppress mean centering and dividing by uncentered second moments instead of variance (\Cref{eq:chan_norm} instead of \Cref{eq:instancenorm}). The corresponding normalization are named with a \papername{} suffix in \Cref{tab:analyzing_mn_tlu} -- for example, GN becomes GFRN, LN becomes LFRN etc. For BN, we just replaced the activation function without changing the normalizing technique, and we observe no significant difference in performance.
We note, however, that IN benefits from use of \momentnorm{} (IN+ReLU vs. \momentnorm{}+ReLU) resulting in 3.61 P@1 gain for ResnetV2-50. Adding \orelu{} leads to another 1.97 points gain (\momentnorm{} + \orelu{}). Similar improvements are observed for InceptionV3. 
In fact, similar improvement trends can be seen for GN and LN as well. This experimental result suggests that \emph{both \momentnorm{} and \orelu{} are critical for the high performance of our method and provide complementary gains}.

\begin{table}[t]
    \centering
    \caption{\textbf{Ablation of our method on Imagenet Classification} for ResnetV2-50~\cite{resnetsv2} and InceptionV3~\cite{szegedy2016rethinking}. We evaluate various combinations of our method with existing normalizations. Combinations that include one of our proposals are marked as \ourcomb{}. Our method, \papername{} + \orelu{}, at the bottom is marked as \ours.
    }
    \vspace{-0.3cm}
    \resizebox{0.85\linewidth}{!}{
    \begin{tabular}{@{}l c c c c@{}}
    \toprule
    \multirow{2}{*}[-2pt]{Method} & \multicolumn{2}{@{}c@{}}{ResnetV2-50} & \multicolumn{2}{@{}c@{}}{InceptionV3} \\
    \cmidrule(lr{0.25em}){2-3} \cmidrule(lr{0.25em}){4-5}
    & P@1 & R@5 & P@1 & R@5 \\
    \midrule
    BN + \relu                   & 76.21 & 92.98 & 78.24 & 94.07 \\
    BN + \orelu{} \ourcomb          & 76.03 & 92.94 & 78.22 & 94.13 \\
    \midrule
    GN + \relu                   & 75.67 & 92.70 & 78.14 & 93.98 \\
    GN + \orelu{} \ourcomb           & 76.59 & 93.16 & 78.50 & 94.18 \\
    GFRN + \relu{} \ourcomb           & 75.93 & 92.65 & 78.16 & 94.03 \\
    GFRN + \orelu{} \ourcomb         & 76.44 & 92.80 & 78.18 & 94.05 \\
    \midrule
    LN + RELU                    & 72.75 & 91.19 & 76.75 & 93.37 \\
    LN + \orelu{}  \ourcomb         & 73.99 & 91.60 & 77.21 & 93.48 \\
    LFRN + RELU \ourcomb             & 75.03 & 92.50 & 77.62 & 93.65 \\
    LFRN + \orelu \ourcomb            & 76.17 & 92.89 & 78.12 & 94.02 \\
    \midrule
    IN + \relu                   & 71.63 & 90.46 & 73.93 & 91.60 \\
    IN + \orelu{} \ourcomb           & 71.72 & 90.53 & 74.81 & 92.01 \\
    FRN + \relu{} \ourcomb            & 75.24 & 92.65 & 77.98 & 94.02 \\
    FRN + \orelu{} \ours          & \textbf{77.21} & \textbf{93.57} & \textbf{78.95} & \textbf{94.49} \\
    \bottomrule
    \end{tabular}
    }
    \label{tab:analyzing_mn_tlu}
    \vspace{-0.4cm}
\end{table}

\medskip
\noindent
\textbf{Models with Fully Connected (FC) layers:}
FC layers are a pathological case for normalization methods, especially for per sample methods (GN, LN, IN, \papername{}), since the number of activations to be normalized over is relatively small. As a result, normalization layers are typically not applied after FC layers. In this section we evaluate the effect of applying normalization after all the layers irrespective of whether they are FC or convolutional layers.
Note that FC layers are the most challenging scenario for \papername{} since we are normalizing over a single activation ($N=1$). We report results for two architectures where the output of FC layers is normalized: 1) InceptionV3 in \Cref{tab:imagenetnormalization} and 2) VGG-A in \Cref{tab:vggaimagenet}. Note that while ResnetV2-50 also has a FC layer after the global pooling to produce logits, normalization is performed before pooling and is thus not relevant here. InceptionV3 has fully connected layers in an auxiliary logits branch while VGG-A has them in the main network. \emph{\papername{} outperforms all other normalization methods even in this challenging scenario on both the architectures}.

Note that, while training InceptionV3 and VGG-A, it was crucial to use learning rate warm-up (refer \Cref{sec:learning_rate}) and learned $\epsilon$ (refer \Cref{sec:learned_eps}) for \papername{} to achieve peak performance. \papername{} underperformed other methods on InceptionV3 and failed to learn entirely on VGG-A without warm-up. Other methods were not significantly affected.
We discovered that without the warm-up phase, the output of max pooling layers grew to very large magnitudes in first few steps. This saturates the normalized activations (see \Cref{fig:lambdaeffect}) and prevents learning due to poor flow of gradients.

Interestingly, for VGG-A, \batchnorm{} performs worse than `No normalization' at the default learning rate of 0.01. In \Cref{tab:vggaimagenet} we also report results for models trained with a higher learning rate of 0.1. A warm-up phase was useful for all the models at this learning rate. However, the `No normalization' model eventually diverges, while \batchnorm{} shows instability in training (even with warm-up) and performs significantly worse than other methods. In contrast, both \papername{} and GN benefit from training at higher learning rate and yield improved performance with \papername{} outperforming GN.

\begin{table}[t]
    \centering
    \caption{\textbf{Model with fully connected layer.} We provide a comparison on Imagenet Classification for the VGG-A model that uses two fully connected layers. 
    Top half shows the results for training with an initial learning rate of 0.01 (the default rate).
    Bottom half shows the results for training with a higher learning rate of 0.1. The base model diverges at this rate, while the model with Batchnorm exhibits instability. \papername{} and Groupnorm train well, with \papername{} outperforming all others.}
    \vspace{-0.3cm}
    \resizebox{0.9\linewidth}{!}{
    \begin{tabular}{@{}l c c c@{}}
    \toprule
    Method & Learning rate & P@1 & R@5 \\
    \midrule
    No normalization & 0.01 & 69.04 & 88.99 \\
    Batchnorm & 0.01 & 67.82 & 88.11 \\
    Groupnorm & 0.01 & 69.35 & 89.12 \\
    \papername{}      & 0.01 & \textbf{70.04} & \textbf{89.42} \\
    \midrule
    No normalization  & 0.1 & Diverged & Diverged \\
    Batchnorm    & 0.1 & 62.61 & 84.56 \\
    Groupnorm    & 0.1 & 69.94 & 89.57 \\
    \papername{} & 0.1 & \textbf{71.66} & \textbf{90.69} \\
    \bottomrule
    \end{tabular}
    }
    \label{tab:vggaimagenet}
    \vspace{-0.4cm}
\end{table}

\medskip
\noindent
\textbf{Comparison of \orelu{} with related variants:}
In \Cref{tab:compare_tlu} we compare \orelu{} with three related variants for ResnetV2-50 on ImageNet. All four correspond to different combinations of having a scale $\kappa$  and bias $\tau$ to compute the threshold. First observe that \orelu{}, despite having a less general form, outperforms others. Second, all variants with a learnable threshold outperform BN, which doesn't benefit from it.  We conclude that a learnable threshold is necessary for high performance in conjunction with \papername{} however it doesn't need to be input dependent.
Interestingly, while two of the variants correspond to commonly known activations -- \relu{} and Parametric \relu{} (PReLU)~\cite{he2015delving}, the third more general form, termed Affine-TLU, outperforms the previous two and has not been explored to the best of our knowledge. Note that Affine-TLU is different from Maxout~\cite{goodfellow2013maxout}, which computes maximum across groups of channels and, unlike Affine-TLU, results in reduced number of channels.

\begin{table}[t]
    \centering
    \caption{\textbf{Comparison of activations} in conjunction with \papername{} on Imagenet Classification for ResnetV2-50. We observe that learnable threshold is key to high performance of our method in comparison to BN, which doesn't benefit from it.}
    \vspace{-0.3cm}
    \resizebox{0.8\linewidth}{!}{
    \begin{tabular}{@{}l c c@{}}
    \toprule
    Method & P@1 & R@5 \\
    \midrule
    BN + $\max(x, 0)$ (\relu{})     & 76.21 & 92.98 \\
    BN + $\max(x, \tau)$ (\orelu{})          & 76.03 & 92.94 \\
    \midrule
    \papername{} + $\max(x, 0)$ (\relu{})                       & 75.24 & 92.65 \\
    $\max(x, \kappa x)$ (PReLU)~\cite{he2015delving}  & 76.43 & 93.30 \\
    $\max(x, \kappa x+ \tau)$ (Affine-TLU)   & 76.71 & 93.32 \\
    $\max(x, \tau)$ (\orelu{})                     & \textbf{77.21} & \textbf{93.57} \\
    \bottomrule
    \end{tabular}
    }
    \label{tab:compare_tlu}
    \vspace{-0.2cm}
\end{table}

\subsection{Object Detection on COCO}\label{sec:det_coco}
Next, we evaluate our method on the task of Object Detection (OD) and demonstrate that it consistently outperforms other normalization methods at all the batch sizes we evaluated on. Since OD frameworks are typically trained with high resolution inputs, they are limited to using small mini-batch sizes. This constraint makes OD an ideal evaluation benchmark for sample based normalization methods that enable training with small batch sizes.

\medskip
\noindent\textbf{Experimental setup.} 
We perform experiments on the COCO dataset~\cite{coco} with 80 object classes. We train using  the \texttt{train2017} set, and evaluate on the 5k images in \texttt{val2017} (\texttt{minival}) split. We report the standard COCO evaluation metrics of mean average precision with different IoU thresholds, namely AP, AP$^{{50}}$, AP$^{{75}}$~\citet{coco}.

\medskip
\noindent
\textbf{Model:} We use the RetinaNet~\cite{lin2017focal} object detection framework. RetinaNet is a unified single stage detector that comprises of three conceptual components: 1) A backbone network, with an off-the-shelf architecture, that acts as a convolutional feature extractor for a given high resolution input image, 2) a convolutional object classification sub-network that acts on the features extracted by the backbone network and, 3) a convolutional bounding box regression sub-network.
We use a ResnetV1-101 Feature Pyramid Network backbone~\cite{lin2017feature} and resize the input images to 1024$\times$1024.

\medskip
\noindent
\textbf{Training:} To simplify experimentation and evaluation, we only compare all methods on models trained from scratch. We justify this choice based on conclusions from~\cite{he2019rethinking} that, by training longer, model trained from scratch can catch up with models trained by fine-tuning pre-trained models. To ensure this, we start with a baseline fine-tuned model, trained by us at the largest batch size 64, that achieves an AP of 38.3 in 25K training steps ($\text{BN}^*$, \Cref{tab:cocodetection}) and is close to the corresponding result of 39.1 reported in~\cite{lin2017focal}. Next, we empirically find the nearest multiple of 25K that achieves similar accuracy when training from scratch to be 125K steps (BN, \Cref{tab:cocodetection}). We set 125K as the base number of training steps for the largest batch size. We train our models using 8 GPUs and experiment with batch sizes in  \{64, 32, 16\} leading to \{8, 4, 2\} images per GPU  respectively. For smaller batch size $M$ we set the training steps $125000{\times}64/M$ and learning rate as $\texttt{base\_lr}{\times}M/64$. We report best performance using $\texttt{base\_lr} \in \{0.01, 0.05, 0.1\}$. All models are trained using a momentum of 0.9 and weight decay of $4{\times}10^{-4}$.

\medskip
\noindent
\textbf{Comparison of normalization methods:} In \Cref{tab:cocodetection} we observe that \papername{} outperforms both \batchnorm{} and GN at all batch sizes, further validating our results in the previous section. In agreement with the observations from \Cref{tab:imagenet_batchsizes} both \papername{} and GN achieve higher accuracy than \batchnorm{} at the evaluated batch sizes. \papername{} outperforms \batchnorm{} by a significant difference of \textbf{0.9} AP points at the largest batch size, and this gap widens to \textbf{8.9} AP points at the smallest batch size. Further, \papername{} consistently achieves higher accuracy than GN.

\begin{table}[t]
\centering
\caption{\textbf{Object detection results on COCO.} Our method, \papername{}, outperforms other methods for all batch sizes. Note that while \batchnorm{} shows a dramatic drop in performance for smaller batch sizes, \papername{} exhibits a comparatively smaller degradation and consistently outperforms GN that also exhibits similarly stable performance. Note that $\text{BN}^*$ models were trained by fine-tuning a imagenet pre-trained model, while others are trained from scratch.}
\vspace{-0.3cm}
\renewcommand{\tabcolsep}{1.mm}
\resizebox{\linewidth}{!}{
\begin{tabular}{@{}l ccc !{\color{gray}\vrule} ccc !{\color{gray}\vrule} ccc@{}}
    \toprule
    Method &
    \multicolumn{3}{@{}c@{}}{AP} &  
    \multicolumn{3}{@{}c@{}}{AP$^{{50}}$} & 
    \multicolumn{3}{@{}c@{}}{AP$^{{75}}$} \\
    \cmidrule(lr{0.25em}){2-10}
    \color{gray}{imgs/gpu} & 8 & 4 & 2 & 8 & 4 & 2 & 8 & 4 & 2\\
    \midrule
    $\text{BN}^*$ & 38.3 & 37.1 & 32.9 & 57.2 & 55.4 & 49.1 & 41.5 & 40.4 & 35.9 \\
    \midrule
    BN            & 38.7 & 37.9 & 30.2 & 56.6 & 55.2 & 44.5 & 42.1 & 41.4 & 32.5 \\
    GN            & 39.3 & 39.0 & 38.7 & 57.8 & 57.5 & 56.9 & 42.6 & 42.3 & 41.8 \\
    \papername{} & \textbf{39.6} & \textbf{39.5} & \textbf{39.1} & \textbf{58.5} & \textbf{58.4} & \textbf{57.5} & \textbf{43.1} & \textbf{43.3} & \textbf{42.3} \\
    \bottomrule
\end{tabular}
}
\label{tab:cocodetection}
\vspace{-0.2cm}
\end{table}

\medskip
\noindent
\textbf{Effect of batch size:} \batchnorm{} exhibits a dramatic degradation in performance, dropping by \textbf{8.5} AP points for the model trained from scratch, as the number of images per GPU is reduced to 2. In comparison, both \papername{} and GN show a relatively more stable accuracy and degrade by less than \textbf{0.6} AP points. Interestingly, the finetuned $\text{BN}^*$ model for the smallest batch size performs 2.7 AP points better than the corresponding BN model trained from scratch, indicating that longer training at this batch size is detrimental to the performance of batchnorm. In contrast, \papername{} maintains a consistent lead for all the metrics across all batch sizes.

\section{Conclusion}

In this paper we proposed the \papername{} layer, a novel combination of Filter Response Normalization (\momentnorm{}) and a Thresholded activation (\orelu{}) function that eliminates the need for batch dependent training. It outperforms \batchnorm{} in a variety of settings and exhibits a consistently high performance in large as well as small batch training. Further, \papername{} also outperforms Group Normalization, a leading sample based normalization alternative to \batchnorm{}, in all the explored settings. We also demonstrated the success of \papername{} in the pathological case of fully connected layers which are typically not normalized. However, since different normalization methods have been successful in different problem domains, e.g. Layer Normalization has been successful in NLP, we leave exploration of these areas with \papername{} as future work.

{\small
\medskip
\noindent\textbf{Acknowledgement.} We would like to thank Vivek Rathod for help with object detection experiments.

\setlength{\bibsep}{0pt}
\bibliographystyle{plainnat}
\bibliography{paper}
}

\end{document}